\pdfoutput=1

\documentclass[11pt]{article}

\usepackage{ACL2023}

\usepackage{times}
\usepackage{latexsym}

\usepackage[T1]{fontenc}

\usepackage[utf8]{inputenc}

\usepackage{microtype}

\usepackage{inconsolata}

\usepackage{bm}
\usepackage{amsmath}
\usepackage{amsfonts}
\usepackage{multirow}
\usepackage{graphicx}
\usepackage{dblfloatfix}        

\usepackage{algorithm}
\usepackage{algorithmicx}
\usepackage[noend]{algpseudocode}

\title{ACTC: Active Threshold Calibration \\ for Cold-Start Knowledge Graph Completion}

\author{Anastasiia Sedova$^{1,2}$ \and Benjamin Roth$^{1,3}$ \\ 
$^1$ Research Group Data Mining and Machine Learning, University of Vienna, Austria \\ 
$^2$ UniVie Doctoral School Computer Science, University of Vienna, Austria \\ 
$^3$ Faculty of Philological and Cultural Studies, University of Vienna, Austria \\ 
\texttt{\{anastasiia.sedova, benjamin.roth\}@univie.ac.at}}

\newcommand{\name}{ACTC }
\newcommand{\algrule}[1][.2pt]{\par\vskip.5\baselineskip\hrule height #1\par\vskip.5\baselineskip}

\begin{document}
\maketitle

\begin{abstract}
Self-supervised knowledge-graph completion (KGC) relies on estimating a scoring model over (entity, relation, entity)-tuples, for example, by embedding an initial knowledge graph.
Prediction quality can be improved by calibrating the scoring model, typically by adjusting the prediction thresholds using manually annotated examples.
In this paper, we attempt for the first time \emph{cold-start} calibration for KGC, where no annotated examples exist initially for calibration, and only a limited number of tuples can be selected for annotation.

Our new method \textbf{ACTC} finds good per-relation thresholds efficiently based on a limited set of annotated tuples.
Additionally to a few annotated tuples, ACTC also leverages unlabeled tuples by estimating their correctness with Logistic Regression or Gaussian Process classifiers.
We also experiment with different methods for selecting candidate tuples for annotation: density-based and random selection.
Experiments with five scoring models and an oracle annotator show an improvement of 7\% points when using ACTC in the challenging setting with an annotation budget of only 10 tuples, and an average improvement of 4\% points over different budgets.



\end{abstract}

\section{Introduction}

\begin{figure}[t!]
\centering
\includegraphics[scale=0.17]{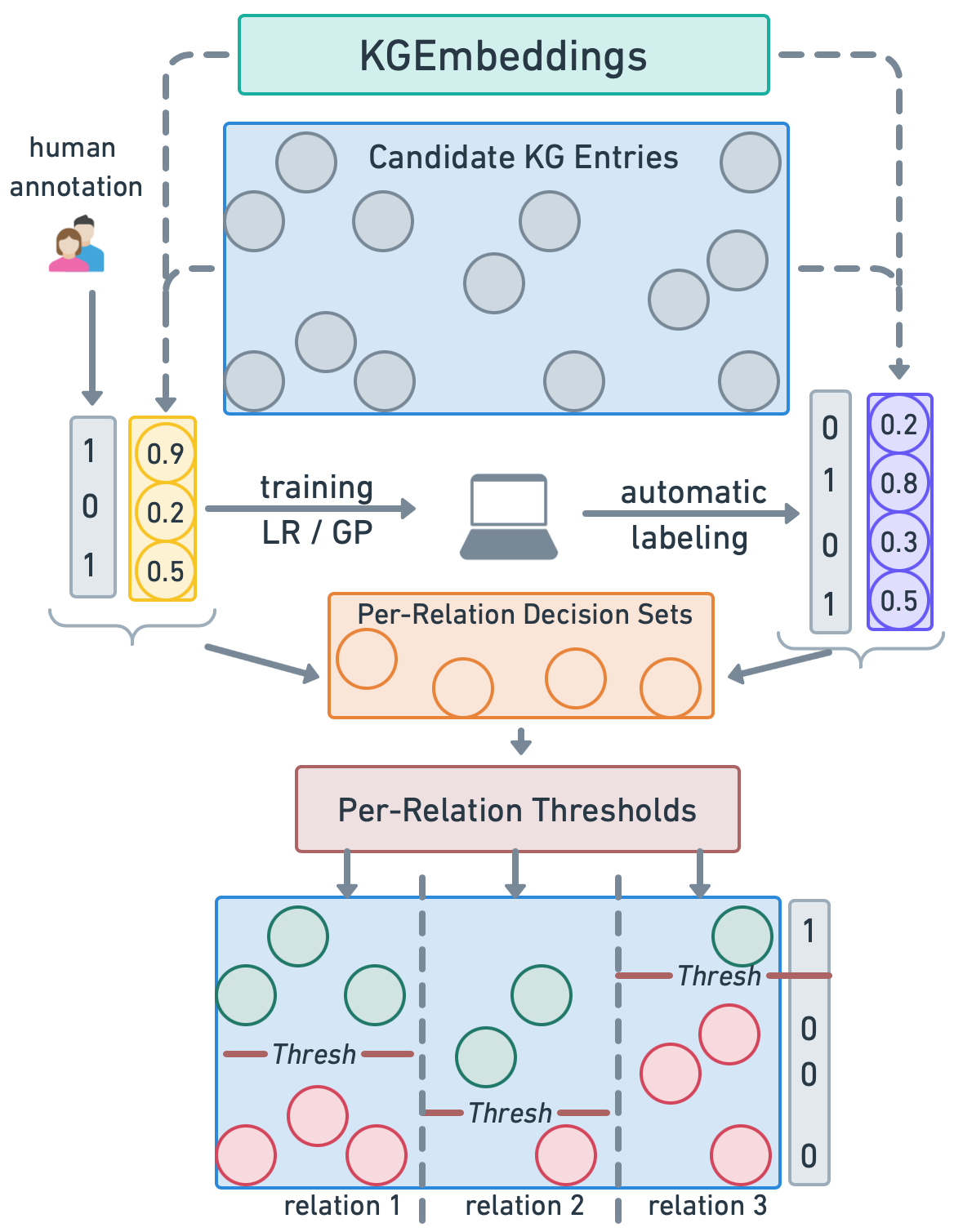}
\setlength{\belowcaptionskip}{-11pt}
\caption{
ACTC method. 
The manually annotated samples are used to train a Logistic Regression or Gaussian Processes classifier, which labels the additional tuples using their scores predicted by a KGE model. 
All annotations (manual and automatic) are later used to estimate the per-relation thresholds.
}
\label{fig:schema}
\end{figure}

Knowledge graphs (KG) organize knowledge about the world as a graph where entities (nodes) are connected by different relations (edges). 
The knowledge-graph completion (KGC) task aims at adding new information in the form of (entity, relation, entity) triples to the knowledge graph.
The main objective is assigning to each triple a \textit{plausibility score}, which defines how likely this triple belongs to the underlying knowledge base.
These scores are usually predicted by the knowledge graph embedding (KGE) models.
However, most KGC approaches do not make any binary decision and provide a ranking, not classification, which does not allow one to use them as-is to populate the KGs \cite{speranskaya}.
To transform the scores into \textit{predictions} (i.e., how probable is it that this triple should be included in the KG), \textit{decision thresholds} need to be estimated.
Then, all triples with a plausibility score above the threshold are classified as positive and included in the KG; the others are predicted to be \textit{negatives} and not added to the KG.
Since the initial KG includes only positive samples and thus cannot be used for threshold calibration, the calibration is usually performed on a manually annotated set of positive and negative tuples (\textit{decision set}).
However, manual annotation is costly and limited, 
and, as most knowledge bases include dozens \cite{tac},
hundreds \cite{toutanova-chen-2015-observed} or even thousands \cite{dbpedia} 
of different relation types, obtaining a sufficient amount of labeled samples for each relation may be challenging. 
This raises a question: 
\newline

\textit{How to efficiently solve the cold-start thresholds calibration problem with \textbf{minimal human input}?}
\newline

We propose a new method for \textbf{Ac}tive \textbf{T}hreshold \textbf{C}alibration \textbf{ACTC}\footnote{The code for ACTC can be found here: https://github.com/anasedova/ACTC}, which estimates the relation thresholds by leveraging unlabeled data additionally to human-annotated data.
In contrast to already existing methods \cite{safavi-koutra-2020-codex, speranskaya} that use only the annotated samples, 
\name labels additional samples automatically with a 
trained predictor 
(Logistic Regression or Gaussian Process model) 
estimated 
on the KGE model scores and available annotations. 
A graphical illustration of ACTC is provided in Figure \ref{fig:schema}. 

Our main contributions are: 
\begin{itemize}
\item  We are the first to study threshold tuning in a budget-constrained environment. 
This setting is more realistic and challenging in contrast to the previous works where large validation sets have been used for threshold estimation. 
\item  We propose actively selecting examples for manual annotation, which is also a novel approach for the KGC setting.
\item  We leverage the unlabeled data to have more labels at a low cost without increasing the annotation budget, which is also a novel approach for the KGC setting.
\end{itemize}

Experiments on several datasets and with different KGE models demonstrate the efficiency of \name for different amounts of available annotated samples, even for as little as one. 





\section{Related Work}
Knowledge graph embedding methods \cite{DBLP:journals/corr/DettmersMSR17, pmlr-v48-trouillon16, NIPS2013_1cecc7a7, 10.5555/3104482.3104584} have been originally evaluated on ranking metrics, not on the actual task of triple classification, which would be necessary for KGC.
More recent works have acknowledged this problem by creating data sets for evaluating KGC (instead of ranking) and proposed simple algorithms for finding prediction thresholds from annotated triples \cite{speranskaya, safavi-koutra-2020-codex}. 
In our work, we study the setting where only a limited amount of such annotations can be provided, experiment with different selection strategies of samples for annotation, and analyze how to use them best.
\citet{10.1145/3308558.3313620} have studied active learning for selecting triples for training a scoring model for KG triples, but their method cannot perform the crucial step of calibration. 
They consequently only evaluate on ranking metrics, not measuring actual link prediction quality.
In contrast, our approach focuses on selecting much fewer samples for optimal \emph{calibration} of a scoring model (using positive, negative, and unlabeled samples).

\section{ACTC: Active Threshold Calibration}
\label{sec:actc}

\name consists of three parts: 
selection of samples for manual annotation,
automatic labeling of additional samples, 
and estimating the per-relation thresholds based on all available labels (manual and automatic ones). 




The first step is selecting unlabeled samples for human annotation. 
In \name this can be done in two ways.
One option is a \textit{random} sampling from the set of all candidate tuples ({$\textbf{ACTC}_{rndm}$}; the pseudocodes can be found in Algorithm \ref{alg:rndm}).
However, not all annotations are equally helpful and informative for estimation.
To select the representative and informative samples that the system can profit the most from, especially with a small annotation budget, we also introduce \textit{density-based} selection $\textbf{ACTC}_{dens}$ inspired by the density-based selective sampling method in active learning \cite{10.1007/978-3-030-58517-4_9, zhu-etal-2008-active} (the pseudocode can be found in Algorithm \ref{alg:dens} in Appendix \ref{app:alg}).
The sample density is measured by summing the squared distances between this sample's score (predicted by the KGE model) and the scores of other samples in the unlabeled dataset.
The samples with the highest density are selected for human annotation.



In a constrained-budget setting with a
limited amount of manual annotations available, there are sometimes only a few samples annotated for some relations and not even one for others.
To mitigate this negative effect and to obtain good thresholds even with limited manual supervision, ACTC labels more samples (in addition to the manual annotations) with a classifier trained on the manually annotated samples to predict the labels based on the KGE model scores.
We experiment with two classifiers: Logistic Regression (\textbf{ACTC-LR}) and Gaussian Processes (\textbf{ACTC-GP}).
The amount of automatically labeled samples depends on hyper-parameter $n$, which reflects the minimal amount of samples needed for estimating each threshold (see ablation study of different \textit{n} values in Section~\ref{sec:ablation}). 
If the number of samples annotated for a relation $r$ ($l_r$) is larger or equal to $n$, only these $l_r$ annotated samples are used for threshold estimation. 
If the amount of manually annotated samples is insufficient (i.e., less than $n$), the additional $n-l_r$ samples are randomly selected from the dataset and labeled by a LR or GP classifier. 
The automatically labeled and manually annotated samples build a per-relation threshold decision set, which contains \textit{at least $n$ samples for a relation $r$ with (manual or predicted) labels}.
The threshold for relation $r$ is later optimized on this decision set.

\begin{algorithm}[t!]
\small
\caption{$ACTC_{rndm}$ algorithm}\label{alg:rndm}
\textbf{Input:} unlabeled dataset $\mathcal{X}$, annotation budget size $l$, minimal decision set size $n$, KGE model $M$, classifier $\mathcal{C}: \mathbb{R} \to [0, 1] $ \\
\textbf{Output}: set of per-relation thresholds $T$
\begin{algorithmic}[1]
\algrule
\Statex \textit{\# Step 1: samples selection for human annotation}
    \State $T \gets$ a set of per-relational thresholds
    \State $\mathcal{X}_{gold} \gets$ randomly selected $l$ samples from $\mathcal{X}$
    \State manually annotate $\mathcal{X}_{gold}$ with $y_{gold}$ labels
    \For {relation $r$}
        \State $\mathcal{X}_{{gold}_{r}} \gets$ samples from $\mathcal{X}_{gold}$ with relation $r$
        \State $y_{{gold}_{r}} \gets$ manual labels for $\mathcal{X}_{{gold}_{r}}$
        \State $scores_{{gold}_{r}} \gets$ KGE model scores for $\mathcal{X}_{{gold}_{r}}$
        \State $l_r \gets |\mathcal{X}_{{gold}_{r}}|$ 
        \Statex \textit{\# Step 2: automatically label additional samples}
        \If{$n > l_r$}
            \State Train a classifier $\mathcal{C}_r$ on  $scores_{{gold}_{r}}$ and $y_{{gold}_{r}}$
            \State $\mathcal{X}_{{auto}_r} \gets$ rand. selected $n-l_r$ samples from $\mathcal{X}$
            \State $scores_{{auto}_{r}} \gets$ KGE model scores for $\mathcal{X}_{{auto}_{r}}$
            \State Predict $y_{{auto}_{r}} = \mathcal{C}_r(scores_{{auto}_{r}})$
            \State $\mathcal{X}_{dec} = (\mathcal{X}_{{gold}_{r}}, y_{{gold}_{r}}) \bigcup (\mathcal{X}_{{auto}_{r}}, y_{{auto}_{r}})$
        \Else
            \State $\mathcal{X}_{dec} = (\mathcal{X}_{{gold}_{r}}, y_{{gold}_{r}}$)
        \EndIf
        \Statex \textit{\# Step 3: estimate per-relation threshold $\tau_r$}
        \State $\tau \gets 0, best\_acc \gets 0$
        \For {$score$ in $scores_{{gold}_{r}}$}
            \State $\tau_i \gets score$
            \State $accuracy_i \gets acc(scores_{{gold}_{r}}, y_{{gold}_{r}} | \tau_i)$
            \If{$accuracy_i > best\_acc$}
                \State $\tau \gets \tau_i$
                \State $best\_acc \gets accuracy_i$
            \EndIf
        \EndFor
        \State $T.append(\tau)$
    \EndFor
\end{algorithmic}
\end{algorithm}

The final part of the algorithm is the estimation of the relation-specific thresholds. 
Each sample score from the decision set is tried out as a potential threshold; the relation-specific thresholds that maximize the local 
accuracy (calculated for this decision set) are selected.


\section{Experiments}

\begin{figure*}[b!]
\centering
\includegraphics[scale=0.373]{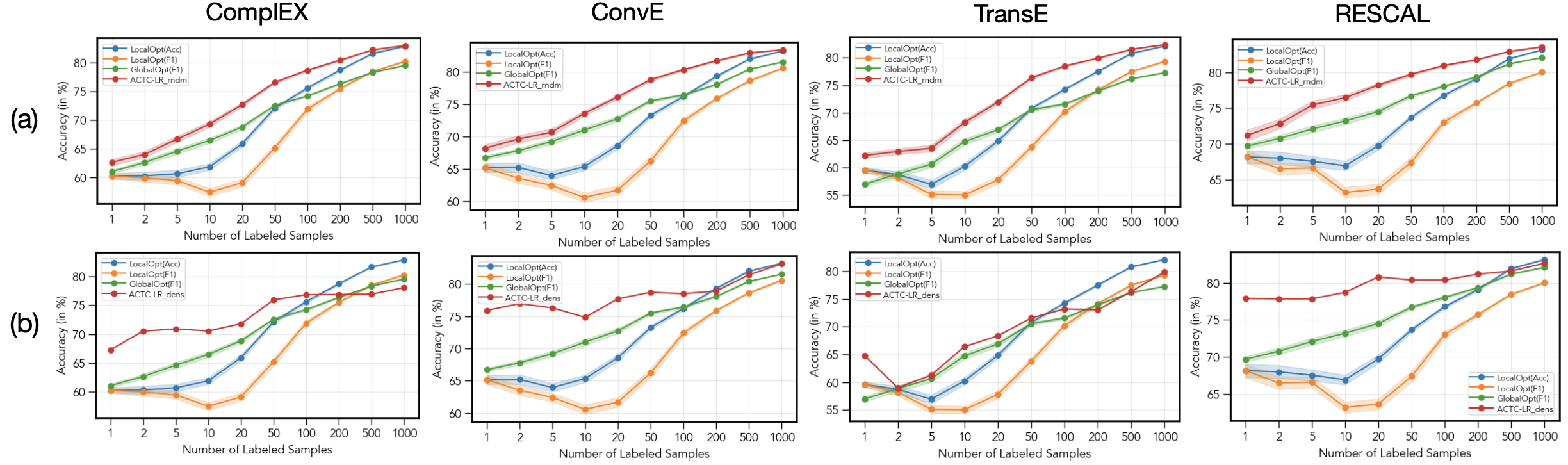}
\vspace*{-5mm}
\setlength{\belowcaptionskip}{-16pt}
\caption{$ACTC-LR_{rndm}$ (upper) and $ACTC-LR_{dens}$ (lower) performance for different amounts of manually annotated samples and with different KGE models.}
\label{fig:lr_rdm_density}
\end{figure*}


We evaluate our method on two KGC benchmark datasets extracted from Wikidata and augmented with manually verified negative samples: CoDEx-s and CoDEx-m\footnote{The third CoDEx dataset, CoDEx-L, is not used in our experiments as it does not provide negative samples.} \cite{safavi-koutra-2020-codex}. 
Some details on their organization are provided in Appendix \ref{sec:data}.
The KGE models are trained on the training sets\footnote{We use the trained models provided by dataset authors.}. 
The ACTC algorithm is applied on the validation sets: 
the gold validation labels are taken as an oracle (\textit{manual annotations}; in an interactive setting they would be presented to human annotators on-the-fly); the remaining samples are used unlabeled. 
The test set is not exploited during ACTC training and serves solely for testing purposes.
The dataset statistics are provided in Table \ref{table:dataset_stats}.
We run our experiments with four KGE models:
ComplEx \cite{pmlr-v48-trouillon16}, ConvE \cite{DBLP:journals/corr/DettmersMSR17},
TransE \cite{NIPS2013_1cecc7a7},
RESCAL \cite{10.5555/3104482.3104584}.
More information is provided in Appendix \ref{sec:models}.

\begin{table}[h!]
\centering
\small
\begin{tabular}
{@{\hspace{0.01cm}}l@{\hspace{0.3cm}}c@{\hspace{0.3cm}}c@{\hspace{0.3cm}}c@{\hspace{0.3cm}}c@{\hspace{0.3cm}}r}
\hline
 Data &  \#Train &  \#Val &  \#Test  &  \#Ent & \#Rel \\
\hline
CoDEx-S &  32,888 & 3,654 &  3,656 & 2,034 & 42 \\
CoDEx-M & 185,584 &  20,620 & 20,622 & 17,050 & 51\\
\hline
\end{tabular}
\setlength{\belowcaptionskip}{-4pt}
\caption{
Datasets statistics.  
The training sets contain only positive triples.
The ratio of positive to negative samples in validation and test sets is 1:1.
}
\label{table:dataset_stats}
\end{table}
\def\arraystretch{0.6}

\begin{table*}[t!]
\centering
\small
\begin{tabular}{@{\hspace{0.08cm}}l@{\hspace{0.08cm}}|c@{\hspace{0.08cm}}c|c@{\hspace{0.08cm}}c|c@{\hspace{0.08cm}}c|c@{\hspace{0.08cm}}c||c@{\hspace{0.08cm}}c|c@{\hspace{0.08cm}}c|c@{\hspace{0.08cm}}c|c@{\hspace{0.08cm}}c@{\hspace{0.08cm}}|c@{\hspace{0.08cm}}c@{\hspace{0.002cm}}c@{\hspace{0.002cm}}}
\\[-0.3em]
\multicolumn{1}{l}{} &
\multicolumn{8}{|c}{CoDEx-s} &
\multicolumn{8}{|c}{CoDEx-m} &
\multicolumn{2}{|c}{Avg}\\
\hline
\\[-0.5em]
\multicolumn{1}{l}{} &
\multicolumn{2}{|c}{ComplEx} & 
\multicolumn{2}{|c}{ConvE} & 
\multicolumn{2}{|c}{TransE} & 
\multicolumn{2}{|c}{RESCAL} & 
\multicolumn{2}{|c}{ComplEx} & 
\multicolumn{2}{|c}{ConvE} & 
\multicolumn{2}{|c}{TransE} & 
\multicolumn{2}{|c|}{RESCAL} \\[0.08em]
\hline
\hline
\\[-0.5em]
& Acc & F1 & Acc & F1 & Acc & F1 & Acc & F1 & Acc & F1 & Acc & F1 & Acc & F1 & Acc & F1 & Acc & F1 \\[0.08em]
\hline
\\[-0.4em]
LocalOpt (Acc) & 70 & 70 & 72 & 72 & 69 & 68 & 74 & 73 & 72 & 70 & 68 & 66 & 65 & 64 & 68 & 67 & 70 & 69\\
\scriptsize \cite{safavi-koutra-2020-codex} & 
\scriptsize$\pm3$ & \scriptsize$\pm3$ & \scriptsize$\pm3$ & \scriptsize$\pm2$ & \scriptsize$\pm3$ & \scriptsize$\pm3$ & \scriptsize$\pm2$ & \scriptsize$\pm2$ & \scriptsize$\pm2$ & \scriptsize$\pm2$ & \scriptsize$\pm3$ & \scriptsize$\pm2$ & \scriptsize$\pm3$ & \scriptsize$\pm3$ & \scriptsize$\pm3$ & \scriptsize$\pm2$ \\[0.6pt]
\multirow{2}{*}{LocalOpt (F1)} & 67 & 69 & 69 & 70 & 65 & 67 & 70 & 71 & 70 & 69 & 66 & 66 & 63 & 64 & 66 & 67 & 67 & 68 \\
& \scriptsize$\pm3$ & \scriptsize$\pm3$ & \scriptsize$\pm3$ & \scriptsize$\pm2$ & \scriptsize$\pm3$ & \scriptsize$\pm3$ & \scriptsize$\pm2$ & \scriptsize$\pm2$ & \scriptsize$\pm2$ & \scriptsize$\pm2$ & \scriptsize$\pm2$ & \scriptsize$\pm2$ & \scriptsize$\pm3$ & \scriptsize$\pm3$ & \scriptsize$\pm3$ & \scriptsize$\pm2$ \\[2pt]
GlobalOpt (F1) & 70 & 74 & 74 & 77 & 68 & 71 & 76 & 79 & 73 & 75 & 68 & 70 & 65 & 68 & 68 & 71 & 70 & 73\\
\scriptsize \cite{speranskaya} & \scriptsize$\pm2$ & \scriptsize$\pm2$ & \scriptsize$\pm1$ & \scriptsize$\pm2$ & \scriptsize$\pm2$ & \scriptsize$\pm2$ & \scriptsize$\pm1$ & \scriptsize$\pm1$ & \scriptsize$\pm1$ & \scriptsize$\pm2$ & \scriptsize$\pm1$ & \scriptsize$\pm2$ & \scriptsize$\pm2$ & \scriptsize$\pm2$ & \scriptsize$\pm1$ & \scriptsize$\pm2$ \\[0.2pt]
\\[-0.5em]
\hline
\hline
\\[-0.5em]
\multirow{2}{*}{$ACTC-LR_{dens}$} & 72 & 72 & \textbf{77} & \textbf{78} & 69 & 71 & 80 & \textbf{81} & \textbf{78} & 77 & \textbf{72} & 71 & 64 & 65 & 72 & 70 & 73 & 73\\
& \scriptsize$\pm3$ & \scriptsize$\pm2$ & \scriptsize$\pm1$ & \scriptsize$\pm1$ & \scriptsize$\pm2$ & \scriptsize$\pm2$ & \scriptsize$\pm1$ & \scriptsize$\pm1$ & \scriptsize$\pm0$ & \scriptsize$\pm1$ & \scriptsize$\pm1$ & \scriptsize$\pm1$ & \scriptsize$\pm1$ & \scriptsize$\pm1$ & \scriptsize$\pm1$ & \scriptsize$\pm1$ \\[1pt]
\multirow{2}{*}{$ACTC-GP_{dens}$} & 72 & 72 & 76 & \textbf{78} & 69 & 71 & 80 & 80 & \textbf{78} & 77 & \textbf{72} & 70 & 64 & 65 & \textbf{73} & 71 & 73 & 73 \\
& \scriptsize$\pm3$ & \scriptsize$\pm2$ & \scriptsize$\pm1$ & \scriptsize$\pm1$ & \scriptsize$\pm1$ & \scriptsize$\pm2$ & \scriptsize$\pm1$ & \scriptsize$\pm1$ & \scriptsize$\pm0$ & \scriptsize$\pm0$ & \scriptsize$\pm1$ & \scriptsize$\pm1$ & \scriptsize$\pm2$ & \scriptsize$\pm2$ & \scriptsize$\pm2$ & \scriptsize$\pm1$ \\[1pt]
\multirow{2}{*}{$ACTC-LR_{rndm}$} & \textbf{74} & \textbf{74} & \textbf{77} & 77 & \textbf{73} & \textbf{72} & 79 & 79 & \textbf{78} & \textbf{78} & \textbf{72} & \textbf{72} & \textbf{69} & \textbf{69} & \textbf{73} & \textbf{73} & \textbf{74} & \textbf{74} \\
& \scriptsize$\pm3$ & \scriptsize$\pm2$ & \scriptsize$\pm2$ & \scriptsize$\pm2$ & \scriptsize$\pm3$ & \scriptsize$\pm3$ & \scriptsize$\pm1$ & \scriptsize$\pm1$ & \scriptsize$\pm1$ & \scriptsize$\pm1$ & \scriptsize$\pm2$ & \scriptsize$\pm2$ & \scriptsize$\pm3$ & \scriptsize$\pm2$ & \scriptsize$\pm2$ & \scriptsize$\pm2$ \\[1pt]
\multirow{2}{*}{$ACTC-GP_{rndm}$} & \textbf{74} & \textbf{74} & \textbf{77} & 77 & \textbf{73} & \textbf{72} & \textbf{81} & \textbf{81} & 77 & 77 & 71 & 71 & 67 & 66 & 72 & 71 & \textbf{74} & \textbf{74} \\
& \scriptsize$\pm3$ & \scriptsize$\pm2$ & \scriptsize$\pm2$ & \scriptsize$\pm2$ & \scriptsize$\pm3$ & \scriptsize$\pm3$ & \scriptsize$\pm1$ & \scriptsize$\pm1$ & \scriptsize$\pm1$ & \scriptsize$\pm1$ & \scriptsize$\pm2$ & \scriptsize$\pm2$ & \scriptsize$\pm3$ & \scriptsize$\pm3$ & \scriptsize$\pm2$ & \scriptsize$\pm2$ \\[1pt]
\hline
\end{tabular}
\setlength{\belowcaptionskip}{-14pt}
\caption{\name results in \% averaged across different sizes of annotation budget reported with the standard error of the mean. 
The experiment with each annotation budget was repeated 100 times. 
}
\label{table:avg_res}
\end{table*}

\def\arraystretch{1}


\subsection{Baselines}

\name is compared to three baselines.
The first baseline \textbf{LocalOpt (Acc)} optimizes the per-relation thresholds towards the accuracy:
for each relation, the threshold is selected from the embedding scores assigned to the samples with manual annotations that contain this relation, so that the \textit{local} accuracy (i.e., accuracy, which is calculated only for these samples) is maximized \citep{safavi-koutra-2020-codex}.
We also modified this approach into \textbf{LocalOpt (F1)} by changing the maximization metric to the local F1 score.
The third baseline is \textbf{GlobalOpt}, where the thresholds are selected by iterative search over a manually defined grid \cite{speranskaya}.
The best thresholds are selected based on the \textit{global} F1 score calculated for the whole dataset\footnote{Labels for samples that include relations for which thresholds have not yet been estimated are calculated using the default threshold of 0.5.}. 
In all baselines, the samples for manual annotation are selected randomly.

\subsection{Results}
We ran the experiments for the following number of manually annotated samples: 1, 2, 5, 10, 20, 50, 100, 200, 500, and 1000.
Experimental setup details are provided in Appendix \ref{sec:exp_set}. 
Table \ref{table:avg_res} provides the result averaging all experiments (here and further, $n=500$ for a fair comparison; see Section \ref{sec:ablation} for analyze of $n$ value), and  
our method ACTC outperforms the baselines in every tried setting as well as on average.
Figure \ref{fig:lr_rdm_density}a also demonstrates the improvement of $ACTC_{rndm}$ over the baselines for every tried amount of manually annotated samples on the example of CoDEx-s dataset; the exact numbers of experiments with different budgets are provided in Appendix \ref{sec:diff_budgets}.
The density-based selection, on the other hand, achieves considerably better results when only few manually annotated samples are available (see Figure \ref{fig:lr_rdm_density}b).
Indeed, choosing representative samples from the highly connected clusters can be especially useful in the case of lacking annotation. 
$LR_{dense}$, which selects points from regions of high density, can be helpful for small annotation budgets since it selects samples that are similar to other samples.
In contrast, when having a sufficient annotation budget and after selecting a certain number of samples, dense regions are already sufficiently covered, and $LR_{rndm}$ provides a more unbiased sample from the entire distribution.





\section{Ablation Study}
\label{sec:ablation}

A more detailed ablation study of different ACTC settings is provided in Appendix \ref{sec:app_ablation}.

\begin{figure}[t!]
\centering
\includegraphics[scale=0.37]{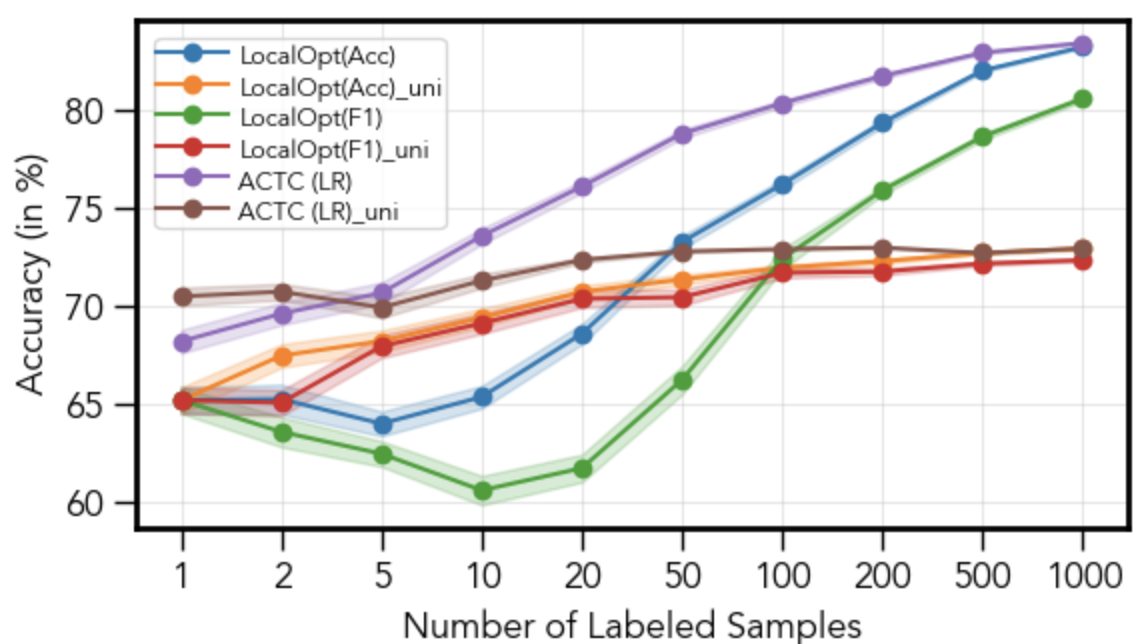}
\vspace*{-1.8mm}
\setlength{\belowcaptionskip}{-10pt}
\caption{The universal threshold calibration methods 
compared to the per-relation methods.}
\label{fig:global}
\end{figure}

\paragraph{Global Thresholds.} 
All methods described above calibrate the \textit{per-relation thresholds}.
Another option is to define a \textit{uniform (uni)} threshold, which works as a generic threshold for all tuples regardless the relations involved.
We implemented it as $ACTC-LR_{uni}$ method, where the additional samples are automatically labeled and used to build a decision dataset together with the manually annotated ones -  in the same way as done for the relation-specific version, but only once for the whole dataset (thus, significantly reducing the computational costs). 
We also applied the LocalOpt(Acc) and LocalOpt(F1) baselines in the uniform setting.
Figure \ref{fig:global} demonstrates the results obtained with the Conve KGE model and random selection mechanism on the CodEX-s dataset.
Although the universal versions generally perform worse than the relation-specific, $ACTC_{uni}$ still outperforms the universal baselines and even relation-specific ones for a small annotation budget. 

\begin{figure}[h!]
\centering
\includegraphics[scale=0.29]{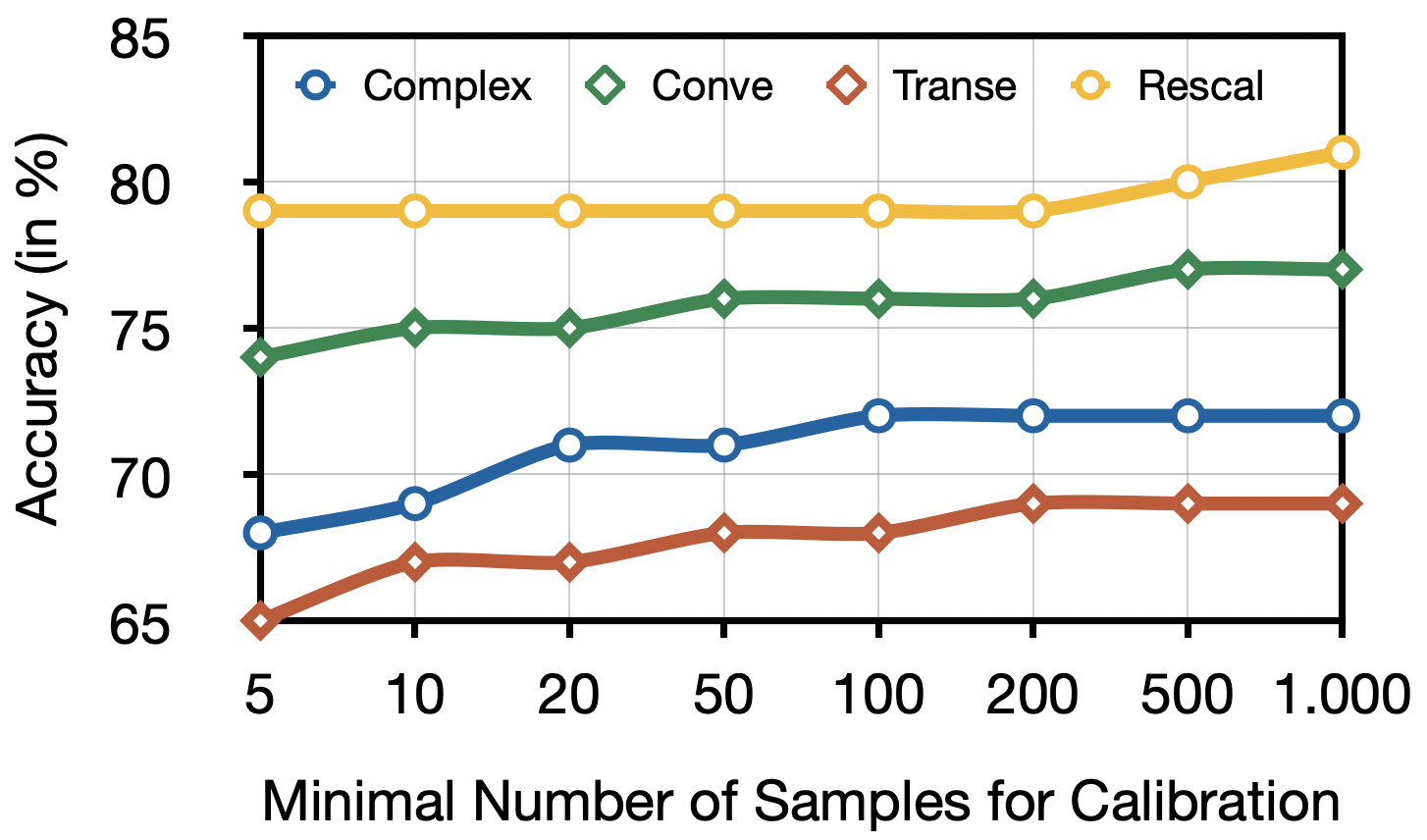}
\setlength{\belowcaptionskip}{-12pt}
\caption{Ablation study for different \textit{n} values. 
}
\label{fig:n_values}
\end{figure}

\paragraph{Different \textit{n} values.}
An important parameter in ACLC is \textit{n},  the minimal sufficient amount of (manually or automatically) labeled samples needed to calibrate the threshold.
The ablation study of different \textit{n} values is provided in Figure \ref{fig:n_values} on the example of $ACTC-LR_{dens}$ setting, averaged across all annotation budgets.
ACTC performs as a quite stable method towards the \textit{n} values.
Even a configuration with a minimum value of $n=5$ outperforms baselines with a small annotation budget or even with quite large one (e.g. for RESCAL). 

\section{Conclusion}
In this work, we explored for the first time the problem of cold-start calibration of scoring models for knowledge graph completion.
Our new method for active threshold calibration ACTC provides different strategies of selecting the samples for manual annotation and automatically labels additional tuples with Logistic Regression and Gaussian Processes classifiers trained on the manually annotated data.
Experiments on datasets with oracle positive and negative triple annotations, and several KGE models, demonstrate the efficiency of our method and the considerable increase in the classification performance even for tiny annotation budgets.
\section{Limitations}


A potential limitation of our experiments is the use of oracle validation labels instead of human manual annotation as in the real-world setting. 
However, all validation sets we used in our experiments were collected based on the manually defined seed set of entities and relations, carefully cleaned and augmented with manually labeled negative samples.
Moreover, 
we chose this more easy-to-implement setting to make our results easily reproducible and comparable with future work. 


Another limitation of experiments that use established data sets and focus on isolated aspects of knowledge-graph construction is their detachment from the real-world scenarios. 
Indeed, in reality knowledge graph completion is done in a much more complicated environment, that involves a variety of stakeholders and aspects, such as data verification, requirements consideration, user management and so on. 
Nevertheless, we do believe that our method, even if studied initially in isolation, can be useful as one component in real world knowledge graph construction. 



\section{Ethics Statement}


Generally, the knowledge graphs used in the experiments are biased towards the North American cultural background, and so are evaluations and predictions made on them. 
As a consequence, the testing that we conducted in our experiments might not reflect the completion performance for other cultural backgrounds.
Due to the high costs of additional oracle annotation, we could not conduct our analysis on more diverse knowledge graphs.
However, we have used the most established and benchmark dataset with calibration annotations, CoDEx, which has been collected with significant human supervision.
That gives us hope that our results will be as reliable and trustworthy as possible. 

 While our method can lead to better and more helpful predictions from knowledge graphs, we cannot guarantee that these predictions are perfect and can be trusted as the sole basis for decision-making, especially in life-critical applications (e.g. healthcare).

\section*{Acknowledgement}
This research
has been funded by the Vienna Science and Technology Fund (WWTF)[10.47379/VRG19008] and
by the Deutsche Forschungsgemeinschaft (DFG, German Research Foundation) RO 5127/2-1.

\bibliography{acl2023}
\bibliographystyle{acl_natbib}

\appendix
\section{$\textbf{ACTC}_{dens}$ Pseudocode}
\label{app:alg}
\def\arraystretch{0.6}

\begin{table*}[b!]
\centering
\small
\begin{tabular}{@{\hspace{0.01cm}}l@{\hspace{0.01cm}}|c@{\hspace{0.01cm}}c|c@{\hspace{0.01cm}}c|c@{\hspace{0.01cm}}c|c@{\hspace{0.01cm}}c||c@{\hspace{0.01cm}}c|c@{\hspace{0.01cm}}c|c@{\hspace{0.01cm}}c|c@{\hspace{0.01cm}}c@{\hspace{0.01cm}}|c@{\hspace{0.01cm}}c@{\hspace{0.01cm}}c@{\hspace{0.01cm}}}
\\[-0.4em]
\multicolumn{1}{l}{} &
\multicolumn{8}{|c}{CoDEx-s} &
\multicolumn{8}{|c}{CoDEx-m} &
\multicolumn{2}{|c}{Avg}\\
\hline
\\[-0.5em]
\multicolumn{1}{l}{} &
\multicolumn{2}{|c}{ComplEx} & 
\multicolumn{2}{|c}{ConvE} & 
\multicolumn{2}{|c}{TransE} & 
\multicolumn{2}{|c}{RESCAL} & 
\multicolumn{2}{|c}{ComplEx} & 
\multicolumn{2}{|c}{ConvE} & 
\multicolumn{2}{|c}{TransE} & 
\multicolumn{2}{|c|}{RESCAL} \\[0.01em]
\hline
\hline
\\[-0.5em]
& Acc & F1 & Acc & F1 & Acc & F1 & Acc & F1 & Acc & F1 & Acc & F1 & Acc & F1 & Acc & F1 & Acc & F1 \\[0.01em]
\hline
\\[-0.4em]
LocalOpt (Acc) & 70 & 70 & 72 & 72 & 69 & 68 & 74 & 73 & 72 & 70 & 68 & 66 & 65 & 64 & 68 & 67 & 70 & 69\\
\scriptsize \cite{safavi-koutra-2020-codex} & 
\scriptsize$\pm3$ & \scriptsize$\pm3$ & \scriptsize$\pm3$ & \scriptsize$\pm2$ & \scriptsize$\pm3$ & \scriptsize$\pm3$ & \scriptsize$\pm2$ & \scriptsize$\pm2$ & \scriptsize$\pm2$ & \scriptsize$\pm2$ & \scriptsize$\pm3$ & \scriptsize$\pm2$ & \scriptsize$\pm3$ & \scriptsize$\pm3$ & \scriptsize$\pm3$ & \scriptsize$\pm2$ \\[0.6pt]
\multirow{2}{*}{LocalOpt (F1)} & 67 & 69 & 69 & 70 & 65 & 67 & 70 & 71 & 70 & 69 & 66 & 66 & 63 & 64 & 66 & 67 & 67 & 68 \\
& \scriptsize$\pm3$ & \scriptsize$\pm3$ & \scriptsize$\pm3$ & \scriptsize$\pm2$ & \scriptsize$\pm3$ & \scriptsize$\pm3$ & \scriptsize$\pm2$ & \scriptsize$\pm2$ & \scriptsize$\pm2$ & \scriptsize$\pm2$ & \scriptsize$\pm2$ & \scriptsize$\pm2$ & \scriptsize$\pm3$ & \scriptsize$\pm3$ & \scriptsize$\pm3$ & \scriptsize$\pm2$ \\[2pt]
GlobalOpt (F1) & 70 & 74 & 74 & 77 & 68 & 71 & 76 & 79 & 73 & 75 & 68 & 70 & 65 & 68 & 68 & 71 & 70 & 73\\
\scriptsize \cite{speranskaya} & \scriptsize$\pm2$ & \scriptsize$\pm2$ & \scriptsize$\pm1$ & \scriptsize$\pm2$ & \scriptsize$\pm2$ & \scriptsize$\pm2$ & \scriptsize$\pm1$ & \scriptsize$\pm1$ & \scriptsize$\pm1$ & \scriptsize$\pm2$ & \scriptsize$\pm1$ & \scriptsize$\pm2$ & \scriptsize$\pm2$ & \scriptsize$\pm2$ & \scriptsize$\pm1$ & \scriptsize$\pm2$ \\[0.2pt]
\\[-0.5em]
\hline
\hline
\\[-0.5em]
\multirow{2}{*}{$ACTC-LR_{dens}$} & 72 & 72 & \textbf{77} & \textbf{78} & 69 & 71 & 80 & \textbf{81} & 78 & 77 & 72 & 71 & 64 & 65 & 72 & 70 & 73 & 73\\
& \scriptsize$\pm3$ & \scriptsize$\pm2$ & \scriptsize$\pm1$ & \scriptsize$\pm1$ & \scriptsize$\pm2$ & \scriptsize$\pm0$ & \scriptsize$\pm1$ & \scriptsize$\pm1$ & \scriptsize$\pm0$ & \scriptsize$\pm1$ & \scriptsize$\pm1$ & \scriptsize$\pm1$ & \scriptsize$\pm1$ & \scriptsize$\pm1$ & \scriptsize$\pm1$ & \scriptsize$\pm1$ \\[1pt]
\multirow{2}{*}{$ACTC-GP_{dens}$} & 72 & 72 & 76 & \textbf{78} & 69 & 71 & 80 & 80 & 78 & 77 & 72 & 70 & 64 & 65 & 73 & 71 & 73 & 73 \\
& \scriptsize$\pm3$ & \scriptsize$\pm2$ & \scriptsize$\pm1$ & \scriptsize$\pm1$ & \scriptsize$\pm1$ & \scriptsize$\pm2$ & \scriptsize$\pm1$ & \scriptsize$\pm1$ & \scriptsize$\pm0$ & \scriptsize$\pm0$ & \scriptsize$\pm1$ & \scriptsize$\pm1$ & \scriptsize$\pm2$ & \scriptsize$\pm2$ & \scriptsize$\pm2$ & \scriptsize$\pm1$ \\[1pt]
\multirow{2}{*}{$ACTC-LR_{rndm}$} & \textbf{74} & \textbf{74} & \textbf{77} & 77 & \textbf{73} & \textbf{72} & 79 & 79 & 78 & 78 & 72 & 72 & 69 & 69 & 73 & 73 & \textbf{74} & \textbf{74} \\
& \scriptsize$\pm3$ & \scriptsize$\pm2$ & \scriptsize$\pm2$ & \scriptsize$\pm2$ & \scriptsize$\pm3$ & \scriptsize$\pm3$ & \scriptsize$\pm1$ & \scriptsize$\pm1$ & \scriptsize$\pm1$ & \scriptsize$\pm1$ & \scriptsize$\pm2$ & \scriptsize$\pm2$ & \scriptsize$\pm3$ & \scriptsize$\pm2$ & \scriptsize$\pm2$ & \scriptsize$\pm2$ \\[1pt]
\multirow{2}{*}{$ACTC-GP_{rndm}$} & \textbf{74} & \textbf{74} & \textbf{77} & 77 & \textbf{73} & \textbf{72} & \textbf{81} & \textbf{81} & 77 & 77 & 71 & 71 & 67 & 66 & 72 & 71 & \textbf{74} & \textbf{74} \\
& \scriptsize$\pm3$ & \scriptsize$\pm2$ & \scriptsize$\pm2$ & \scriptsize$\pm2$ & \scriptsize$\pm3$ & \scriptsize$\pm3$ & \scriptsize$\pm1$ & \scriptsize$\pm1$ & \scriptsize$\pm1$ & \scriptsize$\pm1$ & \scriptsize$\pm2$ & \scriptsize$\pm2$ & \scriptsize$\pm3$ & \scriptsize$\pm3$ & \scriptsize$\pm2$ & \scriptsize$\pm2$ \\[1pt]
\hline
\hline
\\[-0.5em]
\multirow{2}{*}{$ACTC-LR_{dens} (F1)$} & 72 & 72  & 73 & 75 & 63 & 66 & 78 & 79 & 78 & 77 & 72 & 72 & 64 & 66 & 72 & 71 & 72 & 73 \\
& \scriptsize$\pm3$ & \scriptsize$\pm2$ & \scriptsize$\pm0$ & \scriptsize$\pm0$ & \scriptsize$\pm0$ & \scriptsize$\pm1$ & \scriptsize$\pm0$ & \scriptsize$\pm1$ & \scriptsize$\pm1$ & \scriptsize$\pm1$ & \scriptsize$\pm1$ & \scriptsize$\pm2$ & \scriptsize$\pm2$ & \scriptsize$\pm1$ & \scriptsize$\pm3$ & \scriptsize$\pm2$ \\[1pt]
\multirow{2}{*}{$ACTC-GP_{dens} (F1)$} & 72 & 73 & 76 & 78 & 68 & 70 & 79 & 80 & 78 & 77 & 71 & 71 & 64 & 66 & 71 & 73 & 72 & \textbf{74} \\
& \scriptsize$\pm2$ & \scriptsize$\pm1$ & \scriptsize$\pm1$ & \scriptsize$\pm1$ & \scriptsize$\pm1$ & \scriptsize$\pm2$ & \scriptsize$\pm1$ & \scriptsize$\pm1$ & \scriptsize$\pm1$ & \scriptsize$\pm2$ & \scriptsize$\pm2$ & \scriptsize$\pm1$ & \scriptsize$\pm1$ & \scriptsize$\pm1$ & \scriptsize$\pm1$ & \scriptsize$\pm3$ \\[1pt]
\multirow{2}{*}{$ACTC-LR_{rndm} (F1)$} & 73 & 74 & 77 & 78 & 72 & 74 & 79 & 80 & 76 & 75 & 69 & 70 & 66 & 67 & 70 & 70 & 73 & \textbf{74}\\
& \scriptsize$\pm3$ & \scriptsize$\pm2$ & \scriptsize$\pm2$ & \scriptsize$\pm1$ & \scriptsize$\pm3$ & \scriptsize$\pm2$ & \scriptsize$\pm1$ & \scriptsize$\pm1$ & \scriptsize$\pm2$ & \scriptsize$\pm1$ & \scriptsize$\pm2$ & \scriptsize$\pm2$ & \scriptsize$\pm3$ & \scriptsize$\pm2$ & \scriptsize$\pm2$ & \scriptsize$\pm2$ \\[1pt]
\multirow{2}{*}{$ACTC-GP_{rndm} (F1)$} & 74 & 74 & 77 &  77 & 72 & 73 & 79 & 79 & 77 & 77 & 70 & 71 & 67 & 68 & 71 & 72 & 73 & \textbf{74} \\
& \scriptsize$\pm1$ & \scriptsize$\pm2$ & \scriptsize$\pm2$ & \scriptsize$\pm2$ & \scriptsize$\pm3$ & \scriptsize$\pm2$ & \scriptsize$\pm1$ & \scriptsize$\pm1$ & \scriptsize$\pm3$ & \scriptsize$\pm2$ & \scriptsize$\pm1$ & \scriptsize$\pm2$ & \scriptsize$\pm1$ & \scriptsize$\pm3$ & \scriptsize$\pm1$ & \scriptsize$\pm1$ \\[1pt]
\hline
\end{tabular}
\setlength{\belowcaptionskip}{-11pt}
\caption{\name results in \% averaged across different size of annotation budget reported with the standard error of the mean. 
The ACTC method is provided in two local optimization setting: first, the thresholds maximize accuracy (in the same way as it was presented in Figure \ref{table:avg_res}), second, the thresholds are maximized towards F1 score.
The experiment with each annotation budget was repeated 100 times. 
}
\label{table:avg_res_f1}
\end{table*}

\def\arraystretch{1}


\begin{algorithm}[ht]
\small
\caption{$ACTC_{dens}$ algorithm}\label{alg:dens}
\textbf{Input:} unlabeled dataset $\mathcal{X}$, annotation budget size $l$, minimal decision set size $n$, KGE model $M$, classifier $\mathcal{C}: \mathbb{R} \to [0, 1] $ \\
\textbf{Output}: set of per-relation thresholds $T$
\begin{algorithmic}[1]
\algrule
\Statex \textit{\# Step 1: samples selection for human annotation}
    \State $T \gets$ a set of per-relational thresholds
    \For{$i=0, 1, ..., |\mathcal{X}|$}
        \State $density_{x_i} = \sum_{j=0}^{|\mathcal{X}|} (score_j - score_i)^2$
    \EndFor
    \State $\mathcal{X}_{gold} \gets$ top $l$ samples with maximal $density_{x_i}$
    \State manually annotate $\mathcal{X}_{gold}$ with $y_{gold}$ labels
    \State [the rest is the same as in $ACTC_{rndm}$, see Alg. \ref{alg:rndm}] 
        \Statex \textit{\# Step 2: automatically label additional samples}
        \State [same as Step 2 in $ACTC_{rndm}$, see Alg. \ref{alg:rndm}] 
        \Statex \textit{\# Step 3: estimate per-relation threshold $\tau_r$}
        \State [same as Step 3 in $ACTC_{rndm}$, see Alg. \ref{alg:rndm}] 
\end{algorithmic}
\end{algorithm}

\section{CoDEx datasets}
\label{sec:data}
In our experiments, we use benchmark CoDEx datasets \cite{safavi-koutra-2020-codex}.
The datasets were collected based on the Wikidata in the following way: 
a seed set of entities and relations for 13 domains (medicine, science, sport, etc) was defined and used as queries to Wikidata in order to retrieve the entities, relations, and triples. 
After additional postprocessing (e.g. removal of inverse relations), the retrieved data was used to construct 3 datasets: CoDEx-S, CoDEx-M, and CoDEx-L.
For the first two datasets, the authors additionally constructed hard negative samples (by annotating manually the candidate triples which were generated using a pre-trained embedding model), which allows us to use them in our experiments.

\begin{itemize}
    \item An example of positive triple: \textit{(Senegal, part of, West Africa)}.
    \item An example of negative triple: \textit{(Senegal, part of, Middle East)}. 
\end{itemize}

\section{Embedding models}
\label{sec:models}

We use four knowledge graph embedding models. 
This section highlights their main properties and provides their scoring functions. 

\paragraph{ComplEX} \cite{pmlr-v48-trouillon16} uses complex-numbered embeddings and diagonal relation embedding matrix to score triples; 
the scoring function is defined as $s(h, r, t)=\mathbf{e}_{\mathbf{h}}^T \operatorname{diag}\left(\mathbf{r}_{\mathbf{r}}\right) \mathbf{e}_{\mathbf{t}}$. 

\paragraph{ConvE} \cite{DBLP:journals/corr/DettmersMSR17} represents a neural approach to KGE scoring and exploits the non-linearities: $s(h, r, t) = f(\operatorname{vec}(f([\mathbf{e_h} ; \overline{\mathbf{r}}] * \omega)(\mathbf{W}) \mathbf{t}$.

\paragraph{TransE} \cite{NIPS2013_1cecc7a7} is an example of translation KGE models, where the relations are tackled as translations between entities; the embeddings are scored with $s(h, r, t)=-\left\|\mathbf{e}_{\mathbf{h}}+\mathbf{r}_{\mathbf{r}}-\mathbf{e}_{\mathbf{t}}\right\|_p$.

\paragraph{RESCAL} \cite{10.5555/3104482.3104584} treats the entities as vectors and relation types as matrices and scores entities and relation embeddings with the following scoring function:
$s(h, r, t)=\mathbf{e}_{\mathbf{h}}^T \mathbf{R}_{\mathbf{r}} \mathbf{e}_{\mathbf{t}}$. 

These models were selected, first, following the previous works \cite{safavi-koutra-2020-codex, speranskaya}, and, second, to demonstrate the performance of our method using the different KGE approaches: linear (ComplEX and RESCAL), translational (TransE), and neural (ConvE). 

\section{Ablation Study}
\label{sec:app_ablation}


\paragraph{Optimization towards F1 score.}
Just as we converted the LocalOpt (Acc) baseline from \citet{safavi-koutra-2020-codex} to a LocalOpt(F1) setting, we also converted ACTC into ACTC(F1).
The only difference is the metric, which the thresholds maximize: instead of accuracy, the threshold that provides the best F1 scores are looked for. 
Table \ref{table:avg_res_f1} is an extended result table, which provides the ACTC(F1) numbers together with the standard ACTC (optimizing towards accuracy) and baselines.
As can be seen, there is no dramatic change in ACTC performance; naturally enough, the F1 test score for ACTC(F1) experiments is slightly better than the F1 test score for experiments where thresholds were selected based on accuracy value.

\begin{table*}[b!]
\centering
\small
\begin{tabular}{@{\hspace{0.01cm}}l@{\hspace{0.1cm}}|c@{\hspace{0.07cm}}c|c@{\hspace{0.07cm}}c|c@{\hspace{0.07cm}}c|c@{\hspace{0.07cm}}c||c@{\hspace{0.07cm}}c|c@{\hspace{0.07cm}}c|c@{\hspace{0.07cm}}c|c@{\hspace{0.07cm}}c@{\hspace{0.1cm}}|c@{\hspace{0.07cm}}c@{\hspace{0.07cm}}}
\multicolumn{1}{l}{} &
\multicolumn{8}{|c}{CoDEx-s} &
\multicolumn{8}{|c}{CoDEx-m} &
\multicolumn{2}{|c}{Avg} \\
\hline
\multicolumn{1}{l}{} &
\multicolumn{2}{|c}{ComplEx} & 
\multicolumn{2}{|c}{ConvE} & 
\multicolumn{2}{|c}{TransE} & 
\multicolumn{2}{|c}{RESCAL} & 
\multicolumn{2}{|c}{ComplEx} & 
\multicolumn{2}{|c}{ConvE} & 
\multicolumn{2}{|c}{TransE} & 
\multicolumn{2}{|c}{RESCAL} & \\
\hline
\hline
& Acc & F1 & Acc & F1 & Acc & F1 & Acc & F1 & Acc & F1 & Acc & F1 & Acc & F1 & Acc & F1 & Acc & F1\\
\hline
LocalOpt (Acc)$^{1}$ & 60 & 58 & 65 & 65 & 60 & 57 & 68 & 66 & 66 & 63 & 61 & 59 & 55 & 50 & 62 & 58 & 62 & 60 \\
\scriptsize \cite{safavi-koutra-2020-codex} & \scriptsize$\pm1$ & \scriptsize$\pm2$ & \scriptsize$\pm1$ & \scriptsize$\pm1$ & \scriptsize$\pm1$ & \scriptsize$\pm1$ & \scriptsize$\pm1$ & \scriptsize$\pm2$ & \scriptsize$\pm1$ & \scriptsize$\pm2$ & \scriptsize$\pm1$ & \scriptsize$\pm1$ & \scriptsize$\pm0$ & \scriptsize$\pm2$ & \scriptsize$\pm1$ & \scriptsize$\pm2$ &&\\
\multirow{2}{*}{LocalOpt (F1)$^{1}$} & 60 & 58 & 65 & 65 & 60 & 57 & 68 & 66 & 66 & 63 & 61 & 59 & 55 & 50 & 62 & 58 & 62 & 60 \\
& \scriptsize$\pm1$ & \scriptsize$\pm2$ & \scriptsize$\pm1$ & \scriptsize$\pm1$ & \scriptsize$\pm1$ & \scriptsize$\pm1$ & \scriptsize$\pm1$ & \scriptsize$\pm2$ & \scriptsize$\pm1$ & \scriptsize$\pm2$ & \scriptsize$\pm1$ & \scriptsize$\pm1$ & \scriptsize$\pm0$ & \scriptsize$\pm2$ & \scriptsize$\pm1$ & \scriptsize$\pm2$ && \\
GlobalOpt (F1)$^{1}$ & 61 & 67 & 67 & 72 & 57 & 65 & 70 & 75 & 65 & 72 & 60 & 66 & 55 & 62 & 61 & 67 & 62 & 68 \\
\scriptsize \cite{speranskaya}
& \scriptsize$\pm0$ & \scriptsize(1.0) & \scriptsize$\pm0$ & \scriptsize$\pm0$ & \scriptsize$\pm0$ & \scriptsize$\pm1$ & \scriptsize$\pm0$ & \scriptsize$\pm0$ & \scriptsize$\pm1$ & \scriptsize$\pm0$ & \scriptsize$\pm0$ & \scriptsize$\pm0$ & \scriptsize$\pm0$ & \scriptsize(1.0) & \scriptsize$\pm0$ & \scriptsize$\pm0$ && \\
\hline
\hline
\multirow{2}{*}{$ACTC-LR_{dens}^{1}$} & 67 & 68 & 76 & 77 & 65 & 66 & 78 & 79 & 71 & 75 & 69 & 66 & 57 & 65 & 68 & 62 & \textbf{69} & \textbf{70} \\
& \scriptsize$\pm0$ & \scriptsize$\pm0$ & \scriptsize$\pm0$ & \scriptsize$\pm0$ & \scriptsize$\pm0$ & \scriptsize$\pm0$ & \scriptsize$\pm0$ & \scriptsize$\pm0$ & \scriptsize$\pm0$ & \scriptsize$\pm0$ & \scriptsize$\pm0$ & \scriptsize$\pm0$ & \scriptsize$\pm0$ & \scriptsize$\pm0$ & \scriptsize$\pm0$ & \scriptsize$\pm0$ &&\\
\multirow{2}{*}{$ACTC-GP_{dens}^{1}$} & 59 & 47 & 72 & 76 & 61 & 59 & 50 & 67 & 76 & 76 & 71 & 70 & 58 & 66 & 68 & 62 & 64 & 65 \\
& \scriptsize$\pm0$ & \scriptsize$\pm0$ & \scriptsize$\pm0$ & \scriptsize$\pm0$ & \scriptsize$\pm1$ & \scriptsize$\pm1$ & \scriptsize$\pm0$ & \scriptsize$\pm0$ & \scriptsize$\pm0$ & \scriptsize$\pm0$ & \scriptsize$\pm0$ & \scriptsize$\pm0$ & \scriptsize$\pm0$ & \scriptsize$\pm0$ & \scriptsize$\pm0$ & \scriptsize$\pm0$ && \\
\multirow{2}{*}{$ACTC-LR_{rndm}^{1}$} & 63 & 60 & 70 & 69 & 61 & 58 & 76 & 75 & 69 & 67 & 63 & 62 & 57 & 52 & 63 & 60 & 67 & 63  \\
& \scriptsize$\pm0$ & \scriptsize$\pm1$ & \scriptsize$\pm1$ & \scriptsize$\pm1$ & \scriptsize$\pm1$ & \scriptsize$\pm1$ & \scriptsize$\pm1$ & \scriptsize$\pm1$ & \scriptsize$\pm0$ & \scriptsize$\pm2$ & \scriptsize$\pm0$ & \scriptsize$\pm1$ & \scriptsize$\pm0$ & \scriptsize$\pm2$ & \scriptsize$\pm0$ & \scriptsize$\pm1$ && \\
\multirow{2}{*}{$ACTC-GP_{rndm}^{1}$} & 63 & 61& 71 & 70 & 61 & 59 & 76 & 76 & 74 & 73 & 64 & 65 & 56 & 64 & 65 & 64 & 66 & 67 \\
& \scriptsize$\pm1$ & \scriptsize$\pm1$ & \scriptsize$\pm0$ & \scriptsize$\pm1$ & \scriptsize$\pm1$ & \scriptsize$\pm1$ & \scriptsize$\pm1$ & \scriptsize$\pm1$ & \scriptsize$\pm0$ & \scriptsize$\pm0$ & \scriptsize$\pm0$ & \scriptsize$\pm0$ & \scriptsize$\pm0$ & \scriptsize(0.02) & \scriptsize$\pm0$ & \scriptsize$\pm0$ && \\
\hline
\end{tabular}
\caption{ACTC results for $l=1$, $n=500$, averaged across 100 tries for each experiment and reported with the standard error of the mean}
\label{tab:res_1}
\end{table*}

\paragraph{Estimate All Samples} 
Apart from the automatic labeling of \textit{additional} samples discussed in Section~\ref{sec:actc} (i.e., the additional samples are labeled in case of insufficient manual annotations so that the size of the decision set built from manually annotated and automatically labeled samples equals $n$), we also experimented with annotating \textit{all} samples. 
All samples that were not manually labeled are automatically labeled with a classifier. 
However, the performance was slightly better only for the middle budgets (i.e., for the settings with 5, 10, and 20 manually annotated samples) and became considerably worse for large budgets (i.e., 100, 200, etc), especially in the denstity\_selection setting. 
Based on that, we can conclude that a lot of (automatically) labeled additional data is not what the model profits the most; the redundant labels (which are also not gold and potentially contain mistakes) only amplify the errors and lead to worse algorithm performance. 

\paragraph{Hard VS Soft Labels.} 
The classifier's predictions can be either directly used as real-valued \textit{soft} labels or transformed to the \textit{hard} ones by selecting the class with the maximum probability. 
In most of our experiments, the performance of soft and hard labels was practically indistinguishable (yet with a slight advantage of the latter). 
All the results provided in this paper were obtained with hard automatic labels.

\section{Experimental Setting}
\label{sec:exp_set}
As no validation data is available in our setting, the ACTC method does not require any hyperparameter tuning.
We did not use a GPU for our experiments; one ACTC run takes, on average, 2 minutes.
All results are reproducible with a seed value $12345$.

\begin{table*}[t!]
\centering
\small
\begin{tabular}{@{\hspace{0.01cm}}l@{\hspace{0.1cm}}|c@{\hspace{0.1cm}}c|c@{\hspace{0.1cm}}c|c@{\hspace{0.1cm}}c|c@{\hspace{0.1cm}}c||c@{\hspace{0.1cm}}c|c@{\hspace{0.1cm}}c|c@{\hspace{0.1cm}}c|c@{\hspace{0.1cm}}c@{\hspace{0.1cm}}|c@{\hspace{0.1cm}}c@{\hspace{0.1cm}}}
\multicolumn{1}{l}{} &
\multicolumn{8}{|c}{CoDEx-s} &
\multicolumn{8}{|c}{CoDEx-m} &
\multicolumn{2}{|c}{Avg} \\
\hline
\multicolumn{1}{l}{} &
\multicolumn{2}{|c}{ComplEx} & 
\multicolumn{2}{|c}{ConvE} & 
\multicolumn{2}{|c}{TransE} & 
\multicolumn{2}{|c}{RESCAL} & 
\multicolumn{2}{|c}{ComplEx} & 
\multicolumn{2}{|c}{ConvE} & 
\multicolumn{2}{|c}{TransE} & 
\multicolumn{2}{|c}{RESCAL} & \\
\hline
\hline
& Acc & F1 & Acc & F1 & Acc & F1 & Acc & F1 & Acc & F1 & Acc & F1 & Acc & F1 & Acc & F1 & Acc & F1 \\
\hline
LocalOpt (Acc)$^{10}$ & 62 & 62 & 65 & 64 & 60 & 59 & 67 & 66 & 67 & 65 & 63 & 59 & 57 & 57 & 63 & 60 & 63 & 62 \\
\scriptsize \cite{safavi-koutra-2020-codex} &  \scriptsize$\pm1$ & \scriptsize$\pm1$ &  \scriptsize$\pm1$ &  \scriptsize$\pm1$ &  \scriptsize$\pm1$ &  \scriptsize$\pm1$ &  \scriptsize$\pm1$ &  \scriptsize$\pm1$ &  \scriptsize$\pm1$ &  \scriptsize$\pm1$ &  \scriptsize$\pm1$ &  \scriptsize$\pm1$ &  \scriptsize$\pm0$ &  \scriptsize$\pm1$ &  \scriptsize$\pm1$ &  \scriptsize$\pm1$ \\
\multirow{2}{*}{LocalOpt (F1)$^{10}$} & 57 & 61 & 61 & 62 & 55 & 59 & 63 & 64 & 65 & 63 & 60 & 60 & 55 & 58 & 60 & 60 & 60 & 61 \\
&  \scriptsize$\pm1$ &  \scriptsize$\pm1$ &  \scriptsize$\pm1$ &  \scriptsize$\pm1$ &  \scriptsize$\pm1$ &  \scriptsize$\pm1$ &  \scriptsize$\pm1$ &  \scriptsize$\pm1$ &  \scriptsize$\pm1$ &  \scriptsize$\pm1$ &  \scriptsize$\pm1$ &  \scriptsize$\pm1$ &  \scriptsize$\pm1$ &  \scriptsize$\pm1$ &  \scriptsize$\pm1$ &  \scriptsize$\pm1$ \\
GlobalOpt (F1)$^{10}$ & 66 & 71 & 71 & 74 & 65 & 67 & 73 & 76 & 70 & 73 & 65 & 68 & 61 & 66 & 64 & 68 & 67 & 70 \\
\scriptsize \cite{speranskaya} &  \scriptsize$\pm0$ &  \scriptsize$\pm0$ &  \scriptsize$\pm0$ &  \scriptsize$\pm1$ &  \scriptsize$\pm1$ &  \scriptsize$\pm1$ &  \scriptsize$\pm0$ &  \scriptsize$\pm0$ &  \scriptsize$\pm0$ &  \scriptsize$\pm0$ &  \scriptsize$\pm0$ &  \scriptsize$\pm0$ &  \scriptsize$\pm0$ &  \scriptsize$\pm1$ &  \scriptsize$\pm0$ &  \scriptsize$\pm0$ \\
\hline
\hline
\multirow{2}{*}{$ACTC-LR_{dens}^{10}$} & 70 & 73 & 74 & 76 & 66 & 67 & 79 & 80 & 77 & 77 & 71 & 70 & 61 & 61 & 73 & 72 & \textbf{71} & \textbf{72} \\
&  \scriptsize$\pm0$ &  \scriptsize$\pm0$ &  \scriptsize$\pm0$ &  \scriptsize$\pm0$ &  \scriptsize$\pm0$ &  \scriptsize$\pm0$ &  \scriptsize$\pm0$ &  \scriptsize$\pm0$ &  \scriptsize$\pm0$ &  \scriptsize$\pm0$ &  \scriptsize$\pm0$ &  \scriptsize$\pm0$ &  \scriptsize$\pm0$ &  \scriptsize$\pm0$ &  \scriptsize$\pm0$ &  \scriptsize$\pm0$ \\
\multirow{2}{*}{$ACTC-GP_{dens}^{10}$} & 64 & 71 & 73 & 76 & 68 & 65 & 78 & 80 & 77 & 77 & 72 & 70 & 61 & 61 & 73 & 72 & \textbf{71} & \textbf{72} \\
&  \scriptsize$\pm0$ &  \scriptsize$\pm0$ &  \scriptsize$\pm0$ &  \scriptsize$\pm0$ &  \scriptsize$\pm1$ &  \scriptsize$\pm1$ &  \scriptsize$\pm0$ &  \scriptsize$\pm0$ &  \scriptsize$\pm0$ &  \scriptsize$\pm0$ &  \scriptsize$\pm0$ &  \scriptsize$\pm0$ &  \scriptsize$\pm0$ &  \scriptsize$\pm0$ &  \scriptsize$\pm0$ &  \scriptsize$\pm0$ \\
\multirow{2}{*}{$ACTC-LR_{rndm}^{10}$} & 70 & 70 & 74 & 73 & 68 & 66 & 77 & 77 & 74 & 73 & 67 & 66 & 62 & 62 & 67 & 66 & 70 & 70 \\
&  \scriptsize$\pm0$ &  \scriptsize$\pm1$ &  \scriptsize$\pm0$ &  \scriptsize$\pm1$ &  \scriptsize$\pm0$ &  \scriptsize$\pm1$ &  \scriptsize$\pm0$ &  \scriptsize$\pm1$ &  \scriptsize$\pm0$ &  \scriptsize$\pm1$ &  \scriptsize$\pm0$ &  \scriptsize$\pm1$ &  \scriptsize$\pm0$ &  \scriptsize$\pm1$ &  \scriptsize$\pm0$ &  \scriptsize$\pm1$ \\
\multirow{2}{*}{$ACTC-GP_{rndm}^{10}$} & 71 & 70 & 73 & 73 & 68 & 65 & 77 & 77 & 75 & 74 & 68 & 67 & 62 & 61 & 68 & 67 & 70 & 70 \\
&  \scriptsize$\pm0$ &  \scriptsize$\pm1$ &  \scriptsize$\pm0$ &  \scriptsize$\pm1$ &  \scriptsize$\pm1$ &  \scriptsize$\pm1$ &  \scriptsize$\pm0$ &  \scriptsize$\pm1$ &  \scriptsize$\pm0$ &  \scriptsize$\pm1$ &  \scriptsize$\pm0$ &  \scriptsize$\pm1$ &  \scriptsize$\pm1$ &  \scriptsize$\pm1$ &  \scriptsize$\pm1$ &  \scriptsize$\pm1$ \\
\hline
\end{tabular}
\caption{ACTC results for $l=10$, $n=500$, averaged across 100 tries for each experiment and reported with the standard error of the mean}
\label{tab:res_10}
\end{table*}
\def\arraystretch{0.9}

\begin{table*}[t!]
\centering
\small
\begin{tabular}{@{\hspace{0.01cm}}l@{\hspace{0.1cm}}|c@{\hspace{0.15cm}}c|c@{\hspace{0.15cm}}c|c@{\hspace{0.15cm}}c|c@{\hspace{0.15cm}}c||c@{\hspace{0.15cm}}c|c@{\hspace{0.15cm}}c|c@{\hspace{0.15cm}}c|c@{\hspace{0.15cm}}c@{\hspace{0.1cm}}|c@{\hspace{0.1cm}}c@{\hspace{0.1cm}}}
\multicolumn{1}{l}{} &
\multicolumn{8}{|c}{CoDEx-s} &
\multicolumn{8}{|c}{CoDEx-m} & 
\multicolumn{2}{|c}{Avg}\\
\hline
\multicolumn{1}{l}{} &
\multicolumn{2}{|c}{ComplEx} & 
\multicolumn{2}{|c}{ConvE} & 
\multicolumn{2}{|c}{TransE} & 
\multicolumn{2}{|c}{RESCAL} & 
\multicolumn{2}{|c}{ComplEx} & 
\multicolumn{2}{|c}{ConvE} & 
\multicolumn{2}{|c}{TransE} & 
\multicolumn{2}{|c}{RESCAL} & \\
\hline
\hline
& Acc & F1 & Acc & F1 & Acc & F1 & Acc & F1 & Acc & F1 & Acc & F1 & Acc & F1 & Acc & F1 & Acc & F1 \\
\hline
LocalOpt (Acc)$^{50}$ & 72 & 73 & 73 & 74 & 71 & 71 & 74 & 74 & 73 & 72 & 70 & 69 & 67 & 68 & 71 & 70 & 71 & 71 \\
\scriptsize \cite{safavi-koutra-2020-codex} &  \scriptsize$\pm0$ &  \scriptsize$\pm0$ &  \scriptsize$\pm0$ &  \scriptsize$\pm0$ &  \scriptsize$\pm0$ &  \scriptsize$\pm0$ &  \scriptsize$\pm0$ &  \scriptsize$\pm0$ &  \scriptsize$\pm0$ &  \scriptsize$\pm0$ &  \scriptsize$\pm0$ &  \scriptsize$\pm0$ &  \scriptsize$\pm0$ &  \scriptsize$\pm0$ &  \scriptsize$\pm0$ &  \scriptsize$\pm0$ \\
\multirow{2}{*}{LocalOpt (F1)$^{50}$} & 65 & 69 & 66 & 70 & 64 & 68 & 67 & 71 & 69 & 71 & 66 & 68 & 64 & 67 & 67 & 69 & 66 & 69 \\
&  \scriptsize$\pm1$ &  \scriptsize$\pm0$ &  \scriptsize$\pm1$ &  \scriptsize$\pm0$ &  \scriptsize$\pm1$ &  \scriptsize$\pm0$ &  \scriptsize$\pm1$ &  \scriptsize$\pm0$ &  \scriptsize$\pm0$ &  \scriptsize$\pm0$ &  \scriptsize$\pm0$ &  \scriptsize$\pm0$ &  \scriptsize$\pm0$ &  \scriptsize$\pm0$ &  \scriptsize$\pm0$ &  \scriptsize$\pm0$ \\
GlobalOpt (F1)$^{50}$ & 73 & 76 & 75 & 78 & 71 & 73 & 77 & 79 & 74 & 76 & 70 & 72 & 67 & 71 & 69 & 72 & 72 & 75 \\
\scriptsize \cite{speranskaya} &  \scriptsize$\pm0$ &  \scriptsize$\pm0$ &  \scriptsize$\pm0$ &  \scriptsize$\pm0$ &  \scriptsize$\pm0$ &  \scriptsize$\pm0$ &  \scriptsize$\pm0$ &  \scriptsize$\pm0$ &  \scriptsize$\pm0$ &  \scriptsize$\pm0$ &  \scriptsize$\pm0$ &  \scriptsize$\pm0$ &  \scriptsize$\pm0$ &  \scriptsize$\pm0$ &  \scriptsize$\pm0$ &  \scriptsize$\pm0$ \\
\hline
\hline
\multirow{2}{*}{$ACTC-LR_{dens}^{50}$} & 76 & 76 & 78 & 79 & 72 & 72 & 80 & 81 & 79 & 79 & 72 & 72 & 63 & 64 & 73 & 73 & 74 & 75 \\
&  \scriptsize$\pm0$ &  \scriptsize$\pm0$ &  \scriptsize$\pm0$ &  \scriptsize$\pm0$ &  \scriptsize$\pm0$ &  \scriptsize$\pm0$ &  \scriptsize$\pm0$ &  \scriptsize$\pm0$ &  \scriptsize$\pm0$ &  \scriptsize$\pm0$ &  \scriptsize$\pm0$ &  \scriptsize$\pm0$ &  \scriptsize$\pm0$ &  \scriptsize$\pm0$ &  \scriptsize$\pm0$ &  \scriptsize$\pm0$ \\
\multirow{2}{*}{$ACTC-GP_{dens}^{50}$} & 75 & 78 & 78 & 80 & 77 & 78 & 77 & 78 & 79 & 78 & 72 & 72 & 64 & 64 & 74 & 74 & \textbf{75} & 75 \\
&  \scriptsize$\pm0$ &  \scriptsize$\pm0$ &  \scriptsize$\pm0$ &  \scriptsize$\pm0$ &  \scriptsize$\pm0$ &  \scriptsize$\pm0$ &  \scriptsize$\pm0$ &  \scriptsize$\pm0$ &  \scriptsize$\pm0$ &  \scriptsize$\pm0$ &  \scriptsize$\pm0$ &  \scriptsize$\pm0$ &  \scriptsize$\pm0$ &  \scriptsize$\pm0$ &  \scriptsize$\pm0$ &  \scriptsize$\pm0$ \\
\multirow{2}{*}{$ACTC-LR_{rndm}^{10}$} & 76 & 78 & 79 & 79 & 76 & 77 & 80 & 80 & 78 & 78 & 71 & 71 & 69 & 70 & 73 & 73 & \textbf{75} & \textbf{76} \\
&  \scriptsize$\pm0$ &  \scriptsize$\pm0$ &  \scriptsize$\pm0$ &  \scriptsize$\pm0$ &  \scriptsize$\pm0$ &  \scriptsize$\pm0$ &  \scriptsize$\pm0$ &  \scriptsize$\pm0$ &  \scriptsize$\pm0$ &  \scriptsize$\pm0$ &  \scriptsize$\pm0$ &  \scriptsize$\pm0$ &  \scriptsize$\pm0$ &  \scriptsize$\pm0$ &  \scriptsize$\pm0$ &  \scriptsize$\pm0$\\
\multirow{2}{*}{$ACTC-GP_{rndm}^{10}$} & 75 & 78 & 79 & 80 & 77 & 78 & 80 & 80 & 78 & 78 & 72 & 71 & 69 & 70 & 73 & 74 & \textbf{75} & \textbf{76} \\
&  \scriptsize$\pm0$ &  \scriptsize$\pm0$ &  \scriptsize$\pm0$ &  \scriptsize$\pm0$ &  \scriptsize$\pm0$ &  \scriptsize$\pm0$ &  \scriptsize$\pm0$ &  \scriptsize$\pm0$ &  \scriptsize$\pm0$ &  \scriptsize$\pm0$ &  \scriptsize$\pm0$ &  \scriptsize$\pm0$ &  \scriptsize$\pm0$ &  \scriptsize$\pm0$ &
 \scriptsize$\pm0$ &  \scriptsize$\pm0$ \\
\hline
\end{tabular}
\caption{ACTC results for $l=50$, $n=500$, averaged across 100 tries for each experiment and reported with the standard error of the mean}
\label{tab:res_50}
\end{table*}

ACTC does not imply any restrictions on the classifier architecture. 
We experimented with two classifiers: Logistic Regression classifier and Gaussian Processes classifier. 
For both of them, we used a Scikit-learn implementation \cite{pedregosa2011scikit}.
The Logistic Regression classifier was used in the default Scikit-learn setting, with L2 penalty term and inverse of regularization strength equals 100. 
In the Gaussian Processes classifier, we experimented with the following kernels: 
\begin{itemize}
    \item squared exponential \textbf{RBF kernel} with $length\_scale=10$
    \item its generalized and smoothed version, \textbf{Matérn kernel}, with $length\_scale=0.1$
    \item a mixture of different RBF kernels, \textbf{RationalQuadratic} kernel, with $length\_scale=0.1$
\end{itemize}
All the results for Gaussian Processes classifier provided in this paper are obtained with the Matérn kernel \cite{MINASNY2005192} using the following kernel function:

\begin{equation*}
\small
k\left(x_i, x_j\right)=\frac{1}{\Gamma(\nu) 2^{\nu-1}}\left(\frac{\sqrt{2 \nu}}{l} d\left(x_i, x_j\right)\right)^\nu * 
\end{equation*}
\begin{equation*}
\small
* K_\nu\left(\frac{\sqrt{2 \nu}}{l} d\left(x_i, x_j\right)\right)
\end{equation*}

where $K_\nu$ is a Bessel function and $\Gamma$ is a Gamma function.

\section{Results for Different Annotation Budgets}
\label{sec:diff_budgets}
Tables \ref{tab:res_1}, \ref{tab:res_10}, and \ref{tab:res_50} demonstrate the performance of the different ACTC settings for different annotation budgets (1, 10, and 50, respectively).
The results are averaged over all settings; each setting was repeated 100 times. 
Table \ref{tab:res_1} demonstrates how useful and profitable the density-selection methods are in a lower budget setting. 
However, the non-biased random selection works better with more manually annotated samples (e.g., 50).
\end{document}